\documentclass[letterpaper, 10 pt, conference]{ieeeconf}  

\IEEEoverridecommandlockouts                              

\usepackage{amsmath}   

\usepackage{graphicx}
\usepackage{algorithm}
\usepackage{algpseudocode}
\usepackage{setspace}
\usepackage[caption=false,font=normalsize,labelfont=sf,textfont=sf]{subfig}
\usepackage{float}
\usepackage{color}
\usepackage{xcolor}
\usepackage{tabularx,booktabs}
\usepackage{soul}
\newcommand{\rev}[1]{\textcolor{black}{#1}}

\usepackage{ifthen}
\newboolean{revised_version}
\setboolean{revised_version}{false}
\newcommand{\vercmd}[2]{{{#1}}{{#2}}}
\newcommand{\ver}[2]{\ifthenelse {\boolean{revised_version}} {\vercmd{#1}}{\vercmd{#2}}}

\newcolumntype{Y}{>{\centering\arraybackslash}X}

\begin{document}

\title{\LARGE \bf
Functional Eigen-Grasping Using Approach Heatmaps
}

\author{Malek Aburub$^{1}$, Kazuki Higashi$^{2}$, Weiwei Wan$^{1}$ and Kensuke Harada$^{1,3}$
\thanks{$^{1}$ Graduate School of Engineering Science, Osaka University; 1-3 Machikaneyama-cho, Toyonaka, Osaka 560-8531, JAPAN}%
\thanks{$^{2}$ Graduate School of Engineering, Osaka University; 2-1 Yamadaoka, Suita, Osaka 565-0871, Japan}%
\thanks{$^{3}$National Institute of Advanced Industrial Science and Technology, 2-4-7 Aomi, Koto-ku, Tokyo 135-0064, JAPAN}%
\thanks{*Corresponding author e-mail:}
\thanks{ \quad\quad\quad\quad {\tt\small malek@hlab.sys.es.osaka-y.ac.jp}}%
}

\markboth{Journal of \LaTeX\ Class Files,~Vol.~14, No.~8, August~2021}%
{Shell \MakeLowercase{\textit{et al.}}: A Sample Article Using IEEEtran.cls for IEEE Journals}

\maketitle
\thispagestyle{empty}
\pagestyle{empty}

\begin{abstract}

This work presents a framework for a robot with a multi-fingered hand to freely utilize daily tools, including functional parts like buttons and triggers. An approach heatmap is generated by selecting a functional finger, indicating optimal palm positions on the object's surface that enable the functional finger to contact the tool's functional part. Once the palm position is identified through the heatmap, achieving the functional grasp becomes a straightforward process where the fingers stably grasp the object with low-dimensional inputs using the eigengrasp. As our approach does not need human demonstrations, it can easily adapt to various sizes and designs, extending its applicability to different objects. In our approach, we use directional manipulability to obtain the approach heatmap. In addition, we add two kinds of energy functions, i.e., palm energy and functional energy functions, to realize the eigengrasp. Using this method, each robotic gripper can autonomously identify its optimal workspace for functional grasping, extending its applicability to non-anthropomorphic robotic hands. We show that several daily tools like spray, drill, and remotes can be efficiently used by not only an anthropomorphic Shadow hand but also a non-anthropomorphic Barrett hand. 
\end{abstract}

\section{Introduction}

For a robot to work in our everyday environment, it must use our everyday tools, such as sprays, drills, and remote controls. Since such tools include functional parts like triggers and buttons, a robot hand should perform functional grasping, where a robot hand not only grasps the object but also performs the required functions by manipulating the functional part in a post-grasping action. For example, when a robot sprays mist on flowers, a robotic hand may first grasp the bottle and then use one of its fingers to pull the trigger. To realize such functions using a multi-fingered hand, this research aims to provide an efficient and general method of realizing functional grasping. 

Robots are equipped with hands with a variety of kinematic structures. Traditionally, research on functional grasping has been centered on mimicking the human grasp. In \cite{handa2020dexpilot} \cite{qin2022dexmv}, they first obtain the pose of a human hand grasping a tool and then map it to the grasping pose of the robotic hand. Other methods \cite{brahmbhatt2019contactgrasp} \cite{lakshmipathy2022contact} \cite{du2022multi} focus on reproducing the contact surface on the object generated by human grasping demonstrations. These approaches tailored for human hands may not be universally effective for grippers lacking anthropomorphic features. On the other hand, we provide a method allowing both anthropomorphic and non-anthropomorphic hands to realize functional grasping. 


The key idea of realizing the functional grasp is to use the palm position heatmap combined with the eigengrasp \cite{ciocarlie2009hand}. By designating a functional finger to interact with the object, we obtain the approach heatmap, which indicates ideal palm positions on the object's surface. This ensures the functional finger can contact the functional part of the object. The approach heatmap is derived from the directional manipulability of the functional finger.

By referring to this heatmap, we can easily determine the palm's position to achieve the object's associated function. Once the palm position is set, the remaining fingers firmly grasp the object. For planning the grasping pose of these fingers with low-dimensional control inputs, we use eigengrasp, applying three energy functions: palm energy for palm-object contact, contact energy for the remaining fingers, and functional energy for the functional finger.

\begin{figure}[t]
    \centering
    \includegraphics[scale = 0.43]{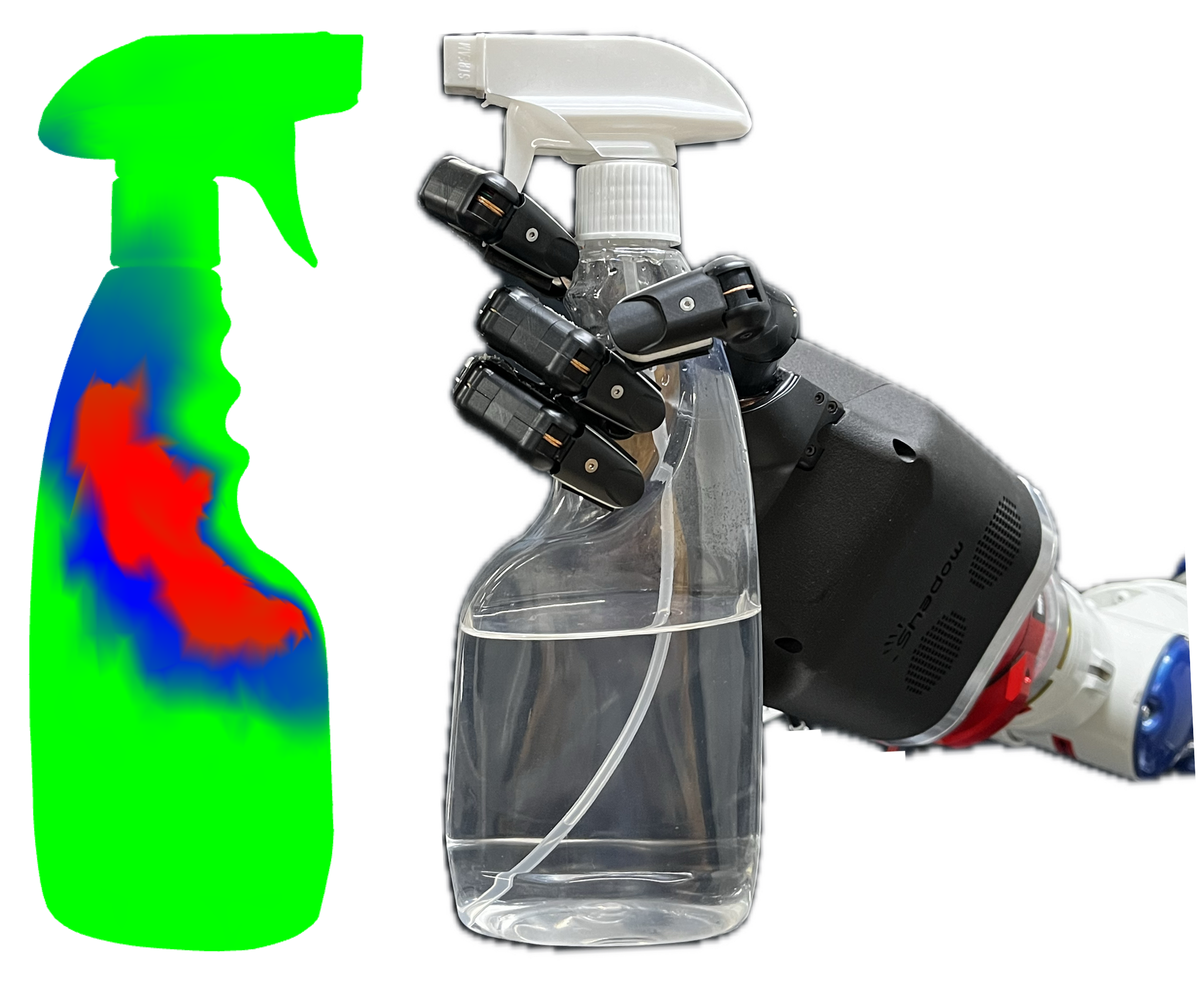}
    \caption{\textbf{Functional grasp}. With the self-generated heatmap on the left and the functional part of the object as input, the grasp planner can generate a grasp capable of satisfying the intended function.}
    \label{grasp_example}
\end{figure}

Here, the only requirement for realizing our proposed approach is that the functional finger has enough DOF to realize the function associated with the object, and the remaining fingers have enough DOF to grasp the object firmly. We show in this paper that our proposed method can be applied not only to anthropomorphic hands but also to non-anthropomorphic hands having less number of DOFs than anthropomorphic ones. We validate the effectiveness of our approach by using the anthropomorphic Shadow hand \cite{kochan2005shadow} and non-anthropomorphic Barret hand \cite{townsend2000barretthand}, where both hands can realize the functional grasp for two kinds of spray bottles, drill, and remote controls. The effectiveness is also experimentally verified by using the Shadow hand. 

To summarize, the main contributions of this work are: 
\begin{itemize}
        
    \item Introduction of novel representation of an approach heatmap for the "palm" of the robotic gripper that takes functionality into consideration
    \item Ease of extension to grippers with lower-dimensional non-anthropomorphic capabilities
    \item Proposal of new energy terms capable of generating functional grasps based on the functional part of the object. 

\end{itemize}

\begin{figure*}[ht]
    \centering
    \includegraphics[width=\linewidth]{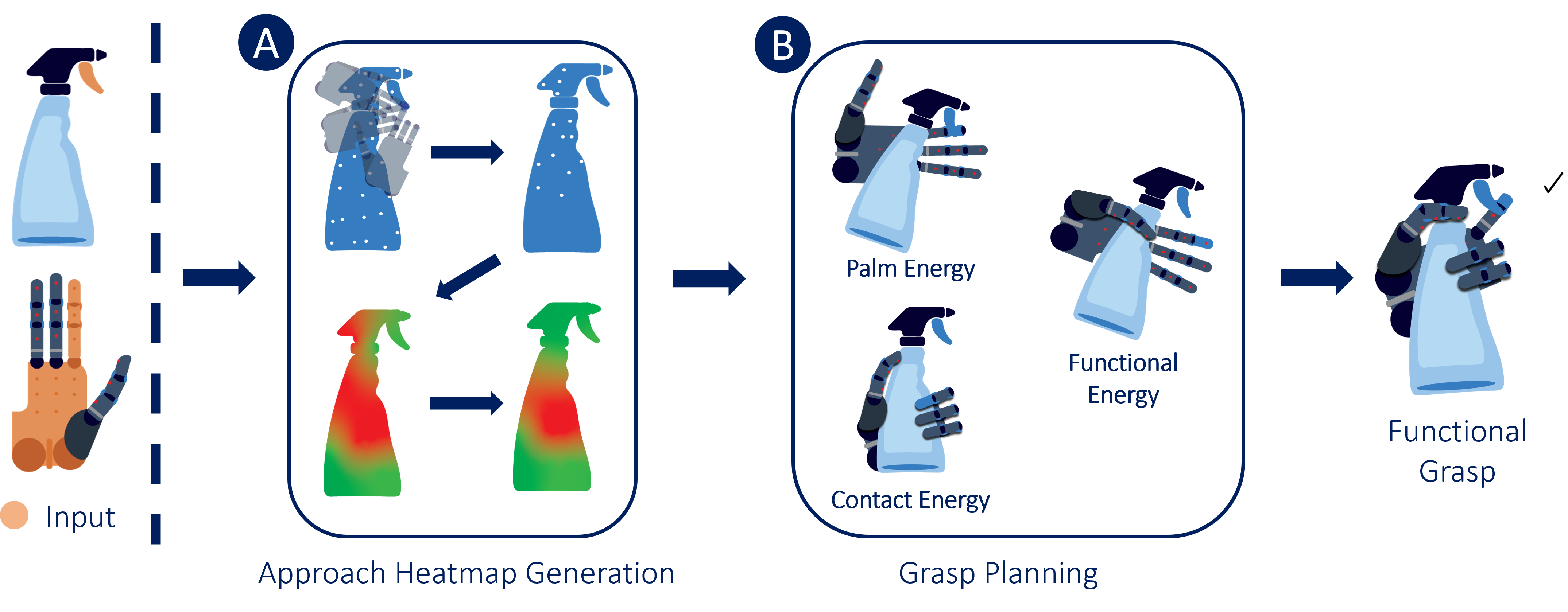}
    \caption{Method Overview - Setting one of the fingers as the functional finger, the approach heatmap is generated in (A) Then we use it in the the grasp planning process using the eigengrasp, applying the energy terms specified in (B) to achieve the functional grasp. }
    \label{overview}
\end{figure*}

\section{Related Works}
Grasp planning is a subject that has been researched extensively over the years \cite{prattichizzo2012manipulability} \cite{rosales2012synthesis} \cite{bicchi2000robotic} where a set of stable grasping poses has been generated by referring to certain quality metrics \cite{nguyen1988constructing} \cite{ferrari1992planning}. The eigengrasp, introduced by Ciocarlie et al. \cite{ciocarlie2009hand}, stands out for its effective synthesis of object grasp poses by narrowing down the search space of the end-effector. Ben Amor et al. \cite{amor2012generalization} create a low-dimensional grasp search space from human demonstrations using a data glove coupled with additional annotations of contact points to optimize the final grasp. Vallery \cite{varley2015generating} proposed replacing human demonstrations with synthetic data from GraspIt! for partial view grasping. However, these methods often overlook the functionality of objects, representing an underexplored area in current research.

On the other hand, functional grasping has only been proposed by a few researchers. By focusing on human-to-robot mapping, Handa et al. \cite{handa2020dexpilot} proposed a teleoperation method for controlling the robotic grippers, \cite{qin2022dexmv} \cite{rajeswaran2017learning} uses learning from demonstration, teaching the robot how to perform a particular task from a few examples of the human performing the same task. 
Ye et al. \cite{ye2023learning} leverage implicit functions and Conditional Variational Autoencoders trained on retargeted human demonstration data.
Nonetheless, these grasp retargeting approaches rely on handcrafted correspondences between the human hand and the robotic gripper, such as joint mapping, fingertip pose retargeting, and hand posture recognition-based mapping. 

ContactGrasp \cite{brahmbhatt2019contactgrasp} generates functional grasp by employing a sample and rank approach, wherein numerous grasps are sampled on an object and filtered based on their resemblance derived from human demonstrations \cite{brahmbhatt2019contactdb}. 
Du et al. \cite{du2022multi} proposed to align the robotic gripper with the human grasp's position and orientation. 
Mandikal \cite{mandikal2021learning} built a model to learn grasp policies that predict the object's grasp affordance based on contact heatmap data. 
Zhang et al. \cite{zhang2023functionalgrasp} proposed a semantic representation of functional hand-object interaction. They could label grasps by annotating specific parts of the hands, guiding their framework to generate functional grasps.
Agarwal et al. \cite{agarwal2023dexterous} used internet data to learn a category-level policy for grasping objects based on their affordance areas.
In all the above methods, the functional grasps have been generated by imitating human grasps.
However, mapping joints and links is not trivial for application to non-anthropomorphic hands.

Hao et al. \cite{dang2014semantic} proposed an example-based grasp planning method. They classify objects, annotate a semantic map with each class's approach direction, hand kinematics, and tactile information, and use it to determine a suitable approach direction for a grasp that satisfies semantic constraints. However, a common limitation among these methods lies in their reliance on human demonstration data or manual annotation, a process that has to be done for every object, which can be time-consuming and hard to apply to grippers of different kinematic structures.

Our method in the approach heatmap generation shares similarities to the work done in reachability analysis \cite{vahrenkamp2012efficient} \cite{akinola2018workspace}, where workspace area is discretized based on reachability. However, our application differs, focusing on the gripper's functional part, employing inverse kinematics and a scoring system. This enables us to uniquely analyze objects for functional grasping, discretizing reachable areas on objects based on functionality and suitability for the intended task.  
In this method, we propose a distinct method for determining the approach direction to target a more intricate grasp where we: 1) eliminate the need for human demonstrations or semantic maps 2) introduce a novel representation of functionality on the object 3) demonstrate compatibility with different grippers.




\section{Method Overview}

In this methodology, we introduce a "functional constraint," ensuring the distal part of the functional finger contacts the functional part of the object. This constraint guides the robotic gripper during object exploration. Our approach to generating functional grasps involves two phases.

In the offline phase, we create an approach heatmap by sampling the object's reachability from the functional meeting point—the intersection of the gripper's functional part and the object. We then solve the inverse kinematics to determine reachability, grading each point based on the directional manipulability index and refining them with a clustering algorithm.

During runtime, the grasp planning process employs eigengrasp. The simulated annealing planner optimizes three energy functions: the palm energy from the approach heatmap, the functional energy for alignment between the gripper and the object, and the contact energy for secure holding of the object. This intricate interplay results in a functional grasp. Figure \ref{overview} provides a visual representation of our method.

\section{Constructing the Approach Heatmap}

Achieving a functional grasp does not only rely on its ability to reach the object but also on its efficiency in carrying out the intended task. To facilitate this, it is necessary to establish grasp affordance regions that guide the robotic gripper in approaching the object. Within this framework, we introduce an approach heatmap to pinpoint the most advantageous positions for the robot to approach the object, ensuring effective task execution. 

This approach draws inspiration from the observed intuition in human grasping. Humans typically approach the functional part of a tool before initiating the grasp, and we leverage this insight by incorporating functional constraint into generating the approach heatmap. This process is illustrated in Fig. \ref{overview}a.

\subsection{Sampling Functional Reachability}

\begin{figure}[ht]
    \centering
    \includegraphics[scale = 0.4]{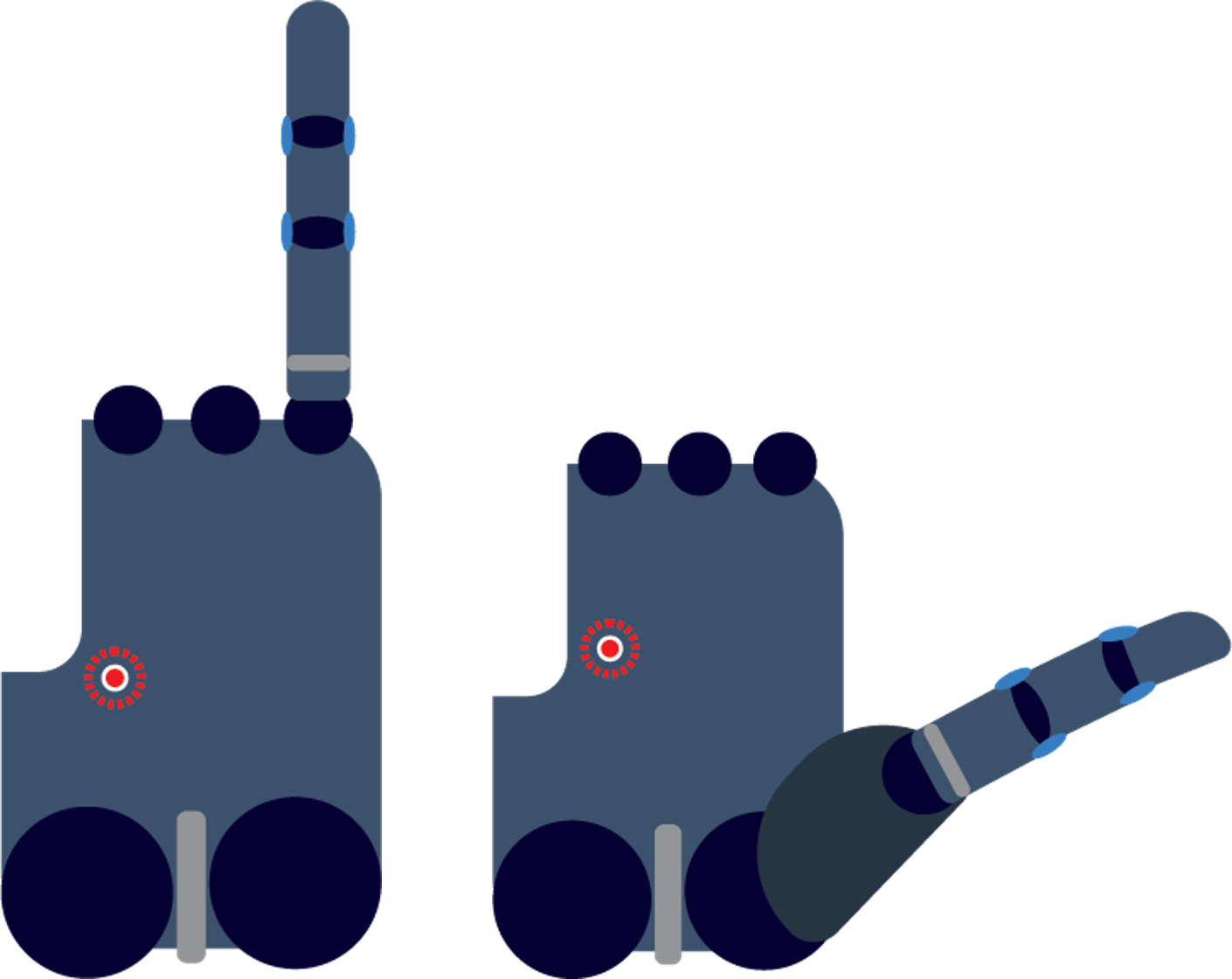}
    \caption{Simplified hand models used for sampling reachability of the object}
    \label{simplified_robot_model}
\end{figure}



To establish the grasp affordance regions, we first identify which object regions the palm can reach while meeting the functional constraint. This is done in a simulated environment by sampling multiple potential approach points on the object's surface and assessing their feasibility for palm access within the functional constraint.

Solving inverse kinematics (IK) for dexterous robotic grippers is complex due to the many joints involved. However, focusing on the reachability of the functional finger to the functional part of the object simplifies the IK problem to a chain of links from the tip of the functional finger to the robotic palm.

For instance, considering the Shadow Hand Lite, as depicted in Fig. \ref{simplified_robot_model}, we set either the thumb or the index to the functional finger, with the palm serving as the end link. At each sampled point, we determine whether a kinematic solution exists for this simplified chain using MoveIt!'s IK solver and retain those instances where successful access is achieved, labeling them as "reachable."

\subsection{Directional Manipulability}

The concept of functional reachability, as it stands, has inherent limitations for two primary reasons. Firstly, it presents a binary classification of either reachable or not, lacking a gradient indicating the transition from non-reachable areas to reachable ones. This binary nature makes it unsuitable for effective use with simulated annealing planners, where providing an energy function is necessary to steer solutions toward more favorable regions. 

Secondly, not all reachable points are conducive to executing the object's intended function. A significant portion of reachable points leads to unfavorable joint configurations for the functional part of the robot, rendering the execution of the intended function challenging.

\begin{algorithm}[ht]
\caption{Approach Heatmap Generation}\label{alg:approachheatmap}
\begin{algorithmic}[1]
\Procedure{Generate Approach Heatmap}{}
\State $points =$ sampleObject()
\State $pointGrade = []$
\For{point in points}
\State $maxGrade = 0$
\For{$T$ iterations} 
\State $reachable =$ computeIk(point)
\If{$reachable$}
\State $score =$ manipulability(jointconfig)
\EndIf
\If{$score > maxGrade$}
\State $maxGrade = score$
\EndIf
\EndFor
\State $pointGrade[point] = [maxGrade]$
\EndFor
\State $Heatmap =$ GenHeatmap(pointGrade)
\EndProcedure
\end{algorithmic}
\end{algorithm}

Therefore, it is necessary to establish a quantification method for each point. In this context, we utilize the directional manipulability index to evaluate each configuration point. The manipulability index, as introduced by Yoshikawa \cite{yoshikawa1985manipulability}, provides a quality value that offers insights into the feasibility of adjustments within the workspace. Visualizing the manipulability ellipsoid offers a geometric representation of the directions where the end-effector can move with minimal or maximal effort. While the condition number indicates maneuverability and proximity to singularity, it proves insufficient for our specific use case. Our goal, such as determining the likelihood of a finger pressing a button, requires more than assessing general movement. Instead of evaluating the maneuverability of the robotic finger in a general sense, by using directional manipulability, we compare the ease of movement relative to the task direction, indicating how likely it is to perform the intended function.

Furthermore, we use the weighted least-norm method mentioned \cite{chan1995weighted} to penalize configurations close to the joint limit. Our analysis centers on the linear component of the resulting Jacobian matrix, from which we extract its eigenvalues and eigenvectors. The number of eigenvalues, denoted as \( m \), corresponds to the dimensions of the Jacobian matrix, which in turn depends on the degrees of freedom of the robot's functional link.

\begin{equation*}
\begin{split}
J_{lin}J_{lin}^T & = V\Lambda V^{-1}\\
                 & = 
\begin{bmatrix}
     &       &       \\
v_{1}& \dots & v_{m} \\
     &       &  
\end{bmatrix}
\begin{bmatrix}
\lambda_{1} & 0      & 0           \\
0           & \ddots & 0           \\
0           & 0      & \lambda_{m}
\end{bmatrix}
 V^{-1}      
\end{split}
\end{equation*}

\begin{figure}[ht]
    \centering
    \includegraphics[scale = 0.3]{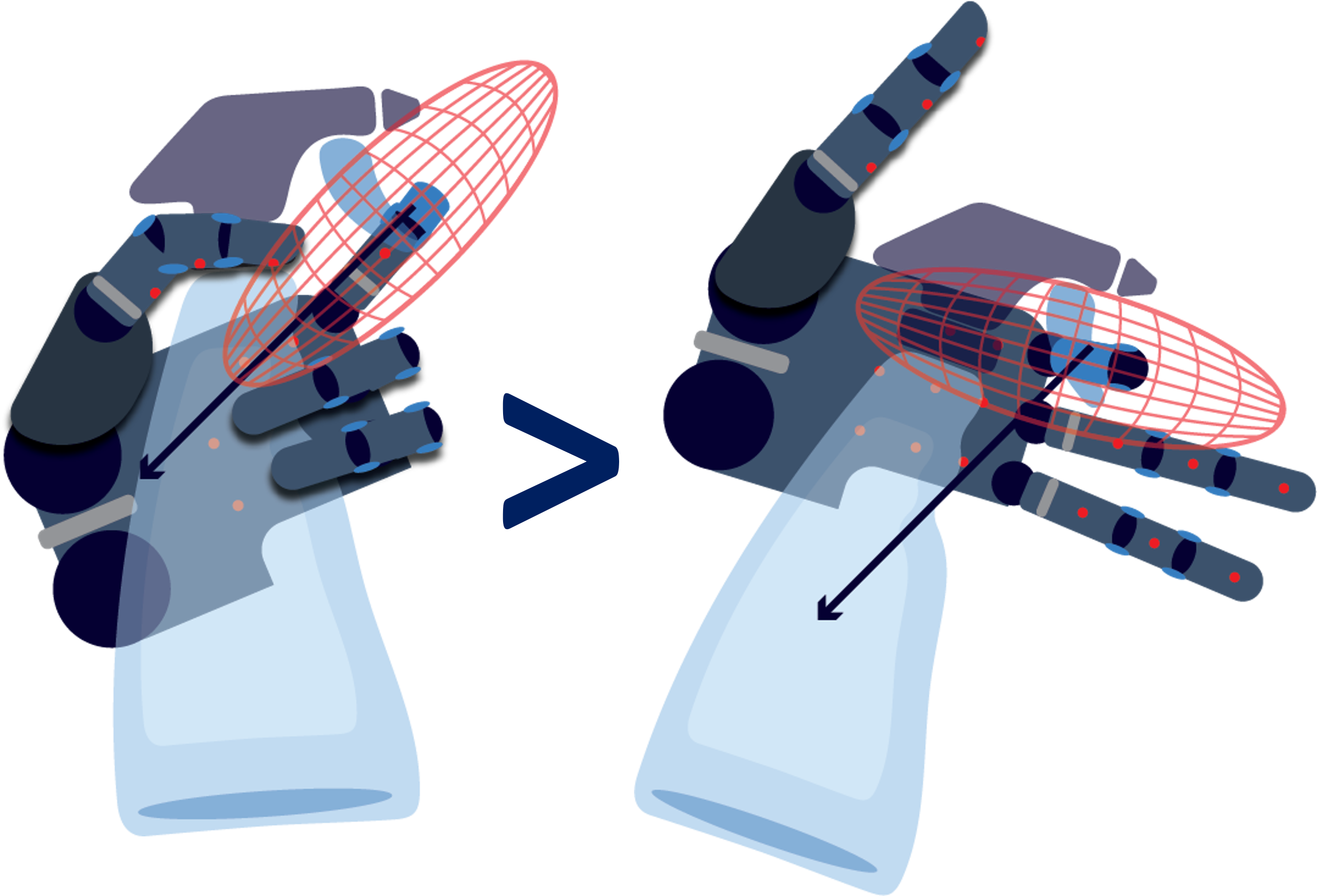}
    \caption{The manipulability ellipsoid of the index finger at different reachable points of the object}
    \label{directional_manipulability}
\end{figure}

This approach enables a comparative analysis of the robot's effectiveness in interacting with each object at different joint configurations. By specifying the function direction \( t \), which is defined along with the functional part (e.g., the rotational joint of a trigger or the direction of button pressing), we can use the equation to grade each point at each joint configuration based on its likelihood to move in that direction, as shown in Fig. \ref{directional_manipulability}.

\[
score = \sum_{i=1}^{m} (t^T.v_i)\lambda_i
\]

The generation of the approach heatmap is a one-time procedure for one object, and Algorithm \ref{alg:approachheatmap} describes this process. In our testing, we divide the object into 500 approach points and evaluate their reachability five times each, prioritizing maximum manipulability. The highest score for each point is recorded, and subsequently, we interpolate these values across the object's mesh. Typically, this entire procedure takes approximately 15 minutes for objects, although this duration can be reduced by decreasing the number of trials or the total number of sampled points.

\subsection{Refining the Approach Heatmap}


The heatmap is refined using the HDBSCAN clustering algorithm \cite{mcinnes2017hdbscan} to remove noise, improving performance by preventing multiple peaks from affecting the grasping algorithm. We select the largest cluster with the highest score on the object, ensuring it represents a significant portion of the heatmap and has a high potential for functional grasping, optimizing overall performance. Fig. \ref{grasp_example} shows this refinement, with the original heatmap in blue and the clustered region in red.

\section{Functional Grasp Planning}

For grasp planning, we use the eigengrasp planner \cite{ciocarlie2009hand} in GraspIt!. This planner effectively narrows the grasp search space to 6 + N dimensions, where 6 represents the parameters for the wrist pose, and N indicates the dimensions to define the hand posture. The grasp planning aims to identify low-energy configurations within this 6 + N dimensional space, representing optimal pose and configurations for a successful grasp. 

To introduce functionality, we incorporate three energy terms: contact, functional, and palm energy terms, where the functional and palm energy terms integrate the functional constraint and the approach heatmap into the grasp planning process. 
Details of each energy function follow in this section. 

\subsection{Contact and Functional Energy}

\begin{figure}[ht]
    \centering
    \includegraphics[scale = 0.3]{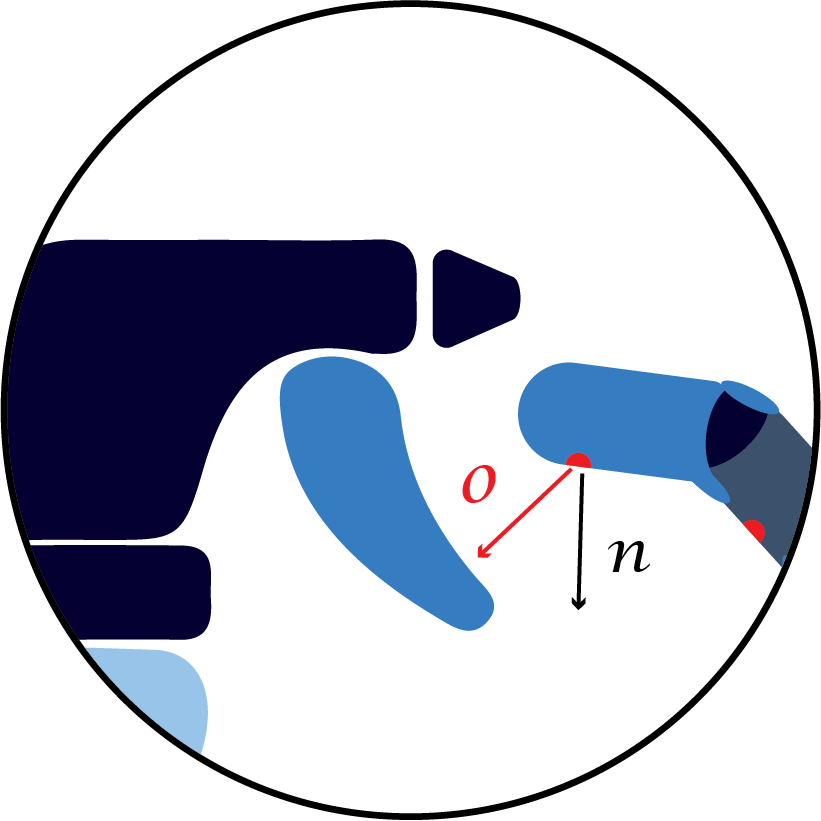}
    \caption{Minimizing the contact energy is done by minimizing the dot product of \textit{n} and \textit{o} while also minimizing the distance \textit{o}}
    \label{contactenergy}
\end{figure}

We implemented the contact energy function from \cite{ciocarlie2009hand}, which evaluates the robotic hand's proximity to the object. As shown in Fig. \ref{contactenergy}, virtual contact points on the gripper measure alignment with the nearest object points by considering both distance and the dot product of vectors between them and the hand's contact normals. This penalizes grasps that are distant or poorly aligned.

\ver{\st{Contact energy is preferred over force closure metrics for achieving functional grasps. While force closure metrics overly focus on the object's center of gravity, leading to suboptimal grasps, contact energy offers a more effective method for reliable and functional grasps.}}{}

Additionally, the functional energy function incorporates the functional constraint into the grasp framework. It operates similarly to the energy function described above but focuses exclusively on the virtual contact points located on the functional part of the robot, assessing their alignment with the functional area of the object. In this case, lower energy function values also indicate better performance.

\subsection{Palm Energy}

\begin{algorithm}[ht]
\caption{Palm Energy Calculation}\label{alg:cap}
\begin{algorithmic}[1]
\Procedure{PalmEnergy}{}
\State $score = 0$
\State $energy = 0$
\State $handPose =$ GetHandTran()
\State $score, distance =$ getScore(handPose)
\If{$score > \omega$} 
\State $palmEnergy = distance -\gamma * score$
\Else 
\State $palmEnergy = Max\_Energy$
\EndIf
\State \Return $palmEnergy$
\EndProcedure
\end{algorithmic}
\label{palmenergyfunction}
\end{algorithm}


This energy function segment, derived from the approach heatmap, establishes a geometric constraint specifying where to grasp the object. As outlined in Algorithm \ref{palmenergyfunction}, it calculates the distance between the robot's palm and the nearest area on the object, considering the area's heatmap score. If the score falls below a threshold $\omega$, it returns the maximum energy value, indicating the region is unsuitable for grasping. Otherwise, it returns the area's average score, adjusted by a factor, gamma, which penalizes moving away from high-score regions. A higher gamma keeps the palm in favorable areas, ensuring proximity to the object and favoring palm poses near high-score regions for lower energy configurations.

\subsection{Hybrid Energy}
The contact, functional, and palm energy functions are blended using the weight parameters $\alpha$, $\beta$, and $\gamma$, ensuring that their combined weight equals 1. These weight parameters determine the degree to which each energy function influences the grasp planning process, as described by the following equation:

\[
E=\alpha E_c + \beta E_f  + \gamma E_p,   
\]
Where $E_c$, $E_f$, and $E_p$ denote the contact, function and palm energy, respectively.

The significance of the contact between the robotic gripper and the object is amplified with higher alpha values, while elevated gamma values anchor the palm more firmly to areas of high score indicated by the approach heatmap. This represents a balance between considering functionality and ensuring a stable, deployment-ready grasp.

\begin{figure*}[ht]
    \centering
    \includegraphics[width=\linewidth]{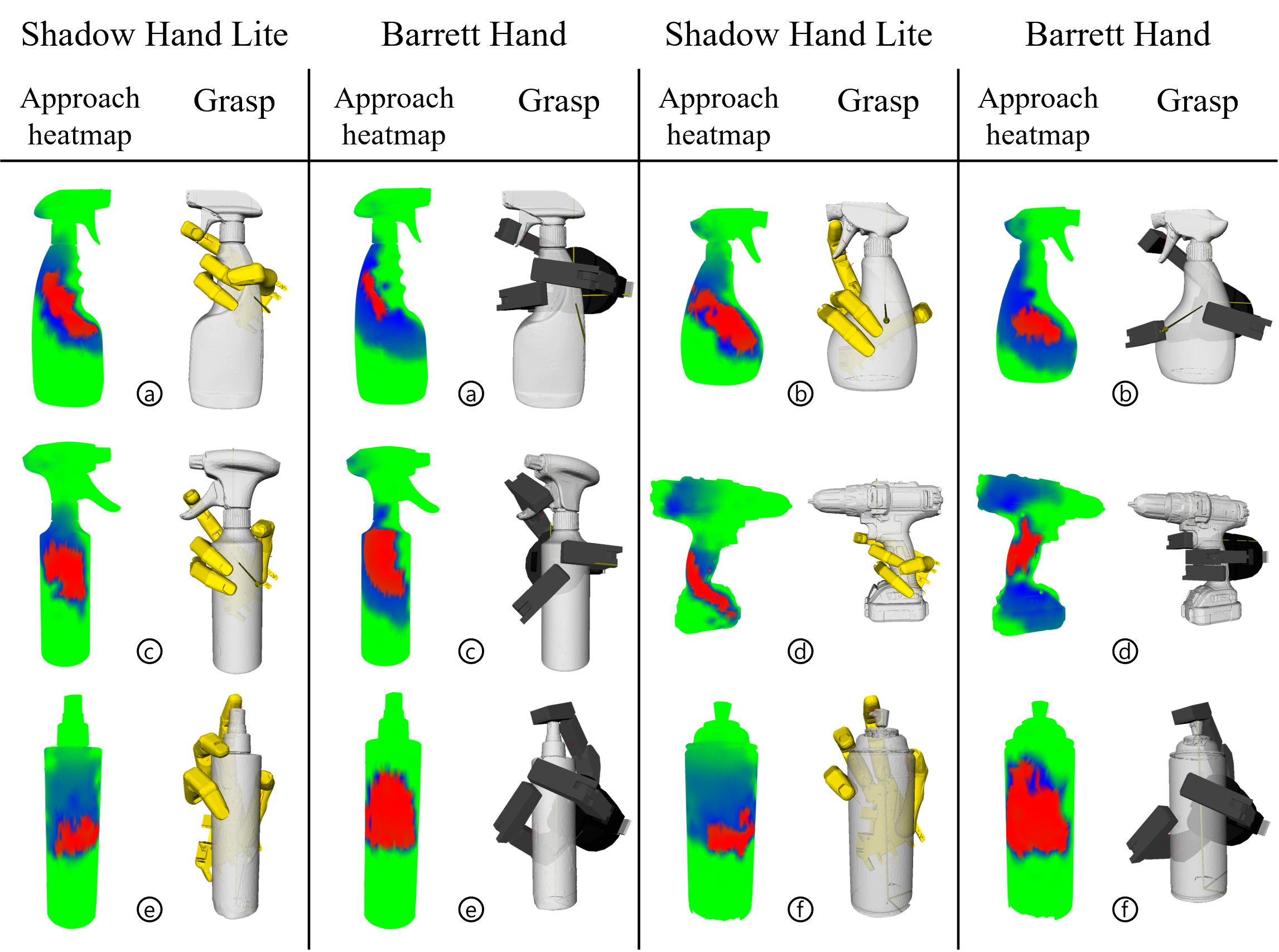}
    \caption{Functional grasps produced for diverse objects with varying designs, each object identified by an alphabet. Each robotic gripper column showcases the approach heatmap on the left, depicting the entire heatmap in blue, the clustered region highlighted in red, and the corresponding functional grasps on the right. All grasps were generated within a 120-second timeframe.}
    \label{allgrasps}
\end{figure*}


\section{Experimental Results}

\subsection{Approach Heatmap Generation}


Approach heatmaps are generated for the anthropomorphic Shadow Hand Lite and the non-anthropomorphic Barrett Hand across the entire object set. 
The functional finger for the Shadow Hand Lite is chosen intuitively to resemble the human hand. The index finger is used for some objects, as shown in Fig. \ref{allgrasps}, while the thumb is selected for others, such as the remote control in Fig. \ref{thumbenergy}.
For the Barrett Hand, the choice is less intuitive. Considering its Degrees of Freedom, one of the coupled fingers is chosen, mimicking the location of the index finger in the human hand.

We did not use a specific object dataset in our experimentation because existing datasets typically lack a variety of functional objects. Instead, we chose a group of everyday objects with multiple designs for testing and generated 3D models of these objects using the Einscan Pro 2X Plus scanner.

Each object was then sampled into 500 potential approach points and evaluated based on weighted directional manipulability. All scores were normalized by dividing them by the 99th quantile score. This one-time offline procedure for each object took approximately 15 minutes on a PC with an Intel Core i7-9750H @ 2.60 GHz CPU.

Fig. \ref{allgrasps} presents distinct approach heatmaps for each object corresponding to different robotic grippers. The entire heatmap is blue, with the clustered region highlighted in red. Notably, among objects serving the same function, variations in design and size influence the differences in the approach heatmap for the respective robotic gripper. Additionally, for different robotic grippers, while certain objects exhibit similarities, there are instances where the high score regions for the Barrett Hand differ from those of the Shadow Hand Lite. This observation underscores the presence of potentially more optimal solutions for robotic grippers lacking similarity to the human hand that are not attainable through a human-to-robot mapping solution. 

\subsection{Planning Functional Grasps}

Using the generated approach heatmaps, the simulated annealing planner of eigengrasp was run for 70,000 steps using the Hybrid energy function, with the lowest energy grasp selected. For the Shadow Hand Lite, the energy function weights were allocated as $\alpha = 0.75$, $\beta = 0.2,$ and $\gamma = 0.05$, while for the Barrett Hand, the weights were $\alpha = 0.6$, $\beta = 0.35$, and $\gamma = 0.05$. 


The differing values of the weights for the Shadow Hand and the Barrett Hand arise from their distinct designs. The Barrett Hand, with its smaller finger size, has a reduced contact weight, making it more challenging to fully envelop the object while still performing the intended function.

\begin{table}[ht]
\centering
\caption{Consistency Rates in GraspIt!}
\begin{tabularx}{0.4\textwidth}{c *{4}{Y}}
\toprule
\multicolumn{1}{c}{Object} & \multicolumn{1}{c}{Shadow Hand Lite} & \multicolumn{1}{c}{Barrett Hand} \\
\midrule
a     & 100\%   & 100\%  \\
b     & 40\%    & 100\%  \\
c     & 100\%   & 80\%   \\
d     & 60\%    & 80\%   \\
e     & 100\%   & 60\%   \\
f     & 80\%    & 100\%  \\
g     & 100\%   & -      \\
\midrule
\textbf{avg}  & 82.86\% & 86.67\%\\
\bottomrule
\end{tabularx}
\label{consistencyrate}
\end{table}

\begin{figure}[h]
    \centering
    \includegraphics[scale = 1]{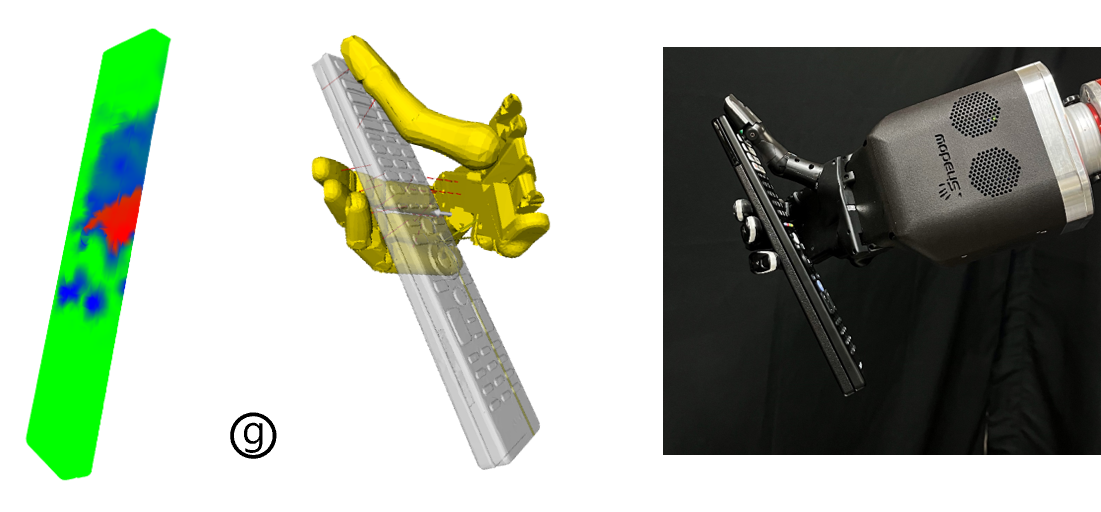}
    \caption{In real-world demonstrations, the thumb acts as the functional finger. For tasks like using a remote control, requiring extensive pre-grasp manipulation, the object is handed to the robotic gripper for a functional grasp.}
    \label{thumbenergy}
\end{figure}

Fig. \ref{allgrasps} presents the experimental results of planning functional grasps, highlighting the functional grasps for each object with corresponding approach heatmaps. Each column represents a robotic gripper and displays the resulting grasp alongside the approach heatmap used during the grasping process. Additionally, we illustrate the use of the thumb as a functional finger in grasping a remote control in Fig. \ref{thumbenergy}, demonstrating the versatility and effectiveness of our approach.

\begin{figure*}[ht]
    \centering
    \includegraphics[width=0.95\linewidth]{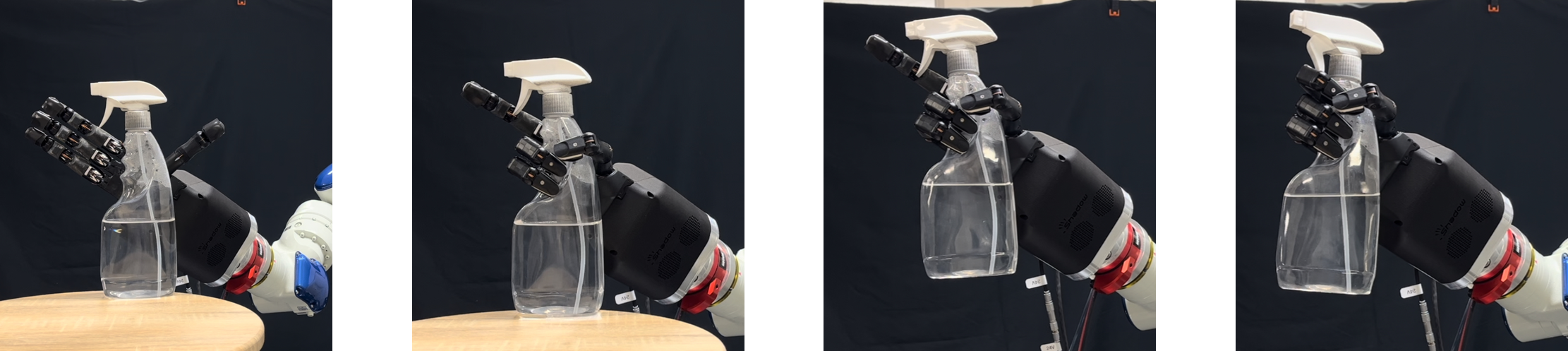}
    \caption{Evaluation Metric: Assess the generated grasps by having the robotic gripper approach and grasp each object using the grasping fingers for stability, then lift the object and use the functional finger to interact with its functional part.}
    \label{lifting_experiment}
\end{figure*}

\begin{figure*}[ht]
    \centering
    \includegraphics[width=0.95\linewidth]{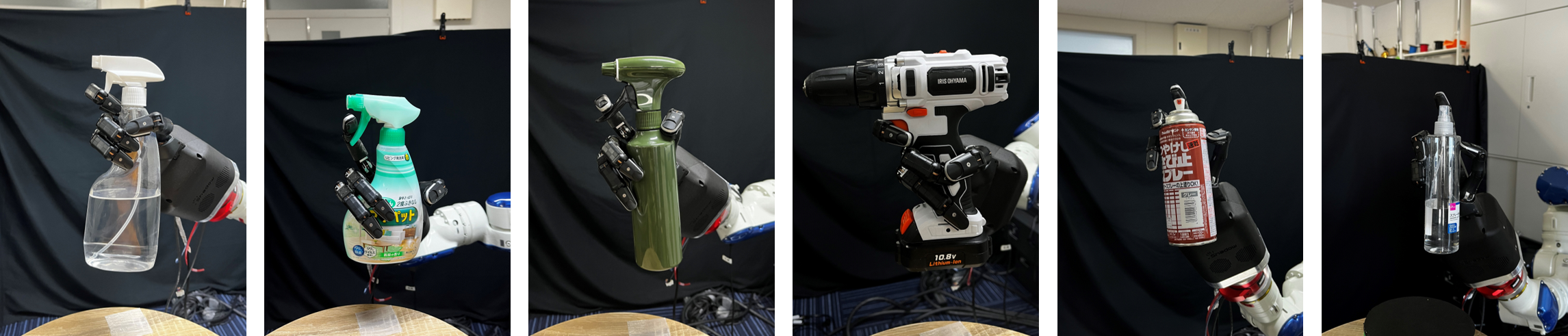}
    \caption{Real World Experiments: Demonstrating functional grasp experiments in real life using the Shadow Hand Lite attached to the Motoman SDA5F robot. }
    \label{real_experiments}
\end{figure*}

To evaluate the consistency of generating functional grasps using approach heatmaps in GraspIt! and to assess the energy functions, we had the grasp planner perform five attempts to generate a functional grasp for each object using the approach heatmap. We then recorded the number of successful grasps from these five attempts.


Table \ref{consistencyrate} summarizes the consistency of each robotic gripper across all objects. Overall, most objects maintained a very high success rate. However, some objects had a lower success rate due to a combination of factors, including object shape, the accessibility of the functional part, and the robot synergy's ability to cover the object effectively. This suggests that running the grasp planner multiple times may be necessary to achieve successful results for these particular objects.

While our method does not rely on any human demonstration or reference, the Shadow Hand Lite's grasps exhibit a human-like resemblance, resulting from anthropomorphic similarities and the alignment of our approach with human intuitiveness. 

It is also important to note that while the method is adaptable to various grippers, certain limitations may arise from the gripper's design. For example, as observed in object e, the Barrett Hand's finger does pose a constraint, being too large for the nozzle spray button, potentially hindering its ability to fulfill the intended function. Additionally, in the remote control case, the buttons proved too small for a functional grasp to be executed effectively by the Barrett Hand.

\subsection{Real World Evaluation}


\begin{table*}[ht]
\centering
\caption{Grasp Resistance to Force Disturbance and Force Amount to Collapse the Grasp}
\begin{tabularx}{\textwidth}{l *{6}{>{\centering\arraybackslash}X}}  
\toprule
\multicolumn{1}{l}{Object} & \multicolumn{1}{c}{a} & \multicolumn{1}{c}{b} & \multicolumn{1}{c}{c} & \multicolumn{1}{c}{d} & \multicolumn{1}{c}{e} & \multicolumn{1}{c}{f} \\
\midrule
Weight (g)                   & 361        & 133     & 243     & 1033    & 148    & 349   \\
Success / \#Trials                     & 30/30        & 30/30     & 30/30     & 30/30    & 30/30    & 16/30   \\
Object Deviation (cm)        & $0.67\pm0.32$    & $0.61\pm0.38$  & $0.89\pm0.49$   & $1.2\pm0.93$    & $0.79\pm0.43$         & $0.52\pm0.21$   \\
Grasp Collapse Force (N)     & $6.02\pm0.51$    & $23.54\pm5$    & $20.76\pm4.1$   & $9.69\pm2.4$    & $4.88\pm0.87$    & $4.7\pm2.3$   \\
\bottomrule
\end{tabularx}
\label{forcelimit}
\end{table*}

Following the planning experiments, we evaluated our grasping framework using the Shadow Hand Lite mounted on a Motoman SDA5F robotic arm. The primary objective was to assess the framework's capability to generate stable and functional grasps in a real-world setting.

Success was determined by the gripper's ability to lift the object steadily, using non-functional fingers for grasping while pressing the designated functional part with the functional finger. Figure \ref{lifting_experiment} demonstrates the gripper successfully pressing the functional button, provided the button's stiffness is manageable.

The Shadow Hand Lite, constrained by its limited force generation, was unable to exert enough force to press the button due to its stiffness. This limitation is discussed in \cite{harada2002sufficient}, suggesting that a robotic gripper with greater force capabilities could potentially resolve this issue.

As shown in Fig. \ref{lifting_experiment}, given the object's pose, the gripper approaches and secures it with the non-functional fingers, lifts it, and presses the functional finger onto the designated area. Fig. \ref{real_experiments} demonstrates the robotic gripper successfully lifting all objects. 
\ver{\st{We evaluated grasp quality using the method in [29] and tested the force needed to break each grasp, up to 30 N. Each object was tested for 10 trials and the results are shown in Table 2; all remained stable except object (f), which was dislodged.}}{}

\rev{We assess the stability of functional grasps, selected based on Table} \ref{consistencyrate}\rev{, using the method from} \cite{kim2013physically}\rev{. Specifically, we apply a 4N disturbance force but instead of relying on human observation for scoring, we quantify grasp stability by measuring the average object deviation caused by the disturbance using markers on the object and an RGB camera. Each grasp is tested in over 30 trials per object, with success defined as the gripper maintaining its hold despite applied disturbances. Additionally, we assess the force required to break each grasp, applying up to 30N. These tests are conducted over 10 trials per object, with results summarized in Table} \ref{forcelimit}\rev{. Among the tested objects, only object F exhibited partial failure, losing its grasp in 14 out of 30 trials.}

\section{Conclusion}

\ver{\st{This paper presents a method for achieving functional grasps across diverse objects using different grippers. By defining the functional components of both the robotic gripper and the object, we integrate a functional constraint to narrow the grasp planning search space, focusing solely on grasps that fulfill the intended function. We introduce the concept of the approach heatmap, which identifies optimal locations on the object for the robotic palm to approach, enabling effective functional grasps. Our experiments demonstrate planning and executing functional grasps on everyday objects using this heatmap.}}{}

\rev{This paper presents a method for achieving functional grasps across diverse objects and grippers by defining functional components of both. We introduce an approach heatmap to identify optimal palm approach locations, narrowing the grasp search space to functionally relevant grasps. Experiments show successful planning and execution on everyday objects using this heatmap.}

\ver{\st{This method allows each robotic gripper to identify suitable grasping areas, accommodating variations in kinematic design and capabilities. Currently, the approach heatmap generation requires the complete model and specification of the object's functional part. To generalize this process to novel and previously unseen objects, we aim to automate it using methods like open-vocabulary part segmentation [30], which can detect the functional part from an RGB image. A key challenge will be integrating the concept of functional reachability to generate the heatmap.\\
Another area for improvement is in integrating the approach heatmap generation and grasp planning processes, as their separation limits the grasp planner's effectiveness, especially for objects like scissors. Future work will focus on integrating these steps to optimize performance and generalize our method, considering the execution of functional motions. We believe this will enhance the generalization and performance of our method.}}{}

\rev{Currently, heatmap generation relies on the object's full model and functional specification. To generalize to unseen objects, we aim to automate this with open-vocabulary part segmentation }\cite{sun2023going}\rev{. A key challenge is integrating functional reachability and grasp planning, which limits effectiveness for complex objects like scissors. Future work will unify these processes to enhance generalization and performance.}



\addtolength{\textheight}{-12cm}   


\bibliographystyle{IEEEtran}
\bibliography{ref}

\end{document}